\documentclass[10pt,twocolumn,letterpaper]{article}

\usepackage[pagenumbers]{cvpr} %

\usepackage{tabularx}
\usepackage{booktabs}

\definecolor{cvprblue}{rgb}{0.21,0.49,0.74}
\usepackage[pagebackref,breaklinks,colorlinks,allcolors=cvprblue]{hyperref}

\def\paperID{*****} %
\def\confName{CVPR}
\def\confYear{2026}

\title{Confidence-Driven Facade Refinement of 3D Building Models Using MLS Point Clouds}

\author{
    Xiaoyu Huang  %
    \\
    Technical University of Munich \\
}

\begin{document}
\maketitle
\begin{abstract}
Digital twins require continuous maintenance to meet the increasing demand for high-precision geospatial data. However, traditional coarse CityGML building models, typically derived from Airborne Laser Scanning (ALS), often exhibit significant geometric deficiencies, particularly regarding facade accuracy due to the nadir perspective of airborne sensors. Integrating these coarse models with high-precision Mobile Laser Scanning (MLS) data is essential to recover detailed facade geometry. Unlike reconstruction-from-scratch approaches that discard existing semantic information and rely heavily on complete data coverage, this work presents an automated refinement framework that utilizes the coarse model as a geometric prior. 
This method enables targeted updates to facade geometry even in complex urban environments. It integrates surface matching to identify outdated surfaces and employs a binary integer optimization to select optimal faces from candidate data. Crucially, hard constraints are enforced within the optimization to ensure the topological validity of the refined output. Experimental results demonstrate that the proposed approach effectively corrects facade misalignments, reducing the Cloud-to-Mesh RMSE by approximately 36\% and achieving centimeter-level alignment. Furthermore, the framework guarantees strictly watertight and manifold geometry, providing a robust solution for upgrading ALS-derived city models.
\end{abstract}
    
\section{Introduction}
\label{sec:intro}

With the development of scanning technologies, point cloud data can be efficiently acquired from diverse platforms, including terrestrial, mobile, and aerial systems \citep{stilla2023change}. This rich data source has become fundamental across various fields, such as autonomous driving \citep{cui2021deep,li2020deep}, robotic navigation \citep{pomerleau2015review,liu2015robotic}, and particularly for the construction of high-precision 3D urban building models \citep{biljecki2015applications, musialski2013survey, haala2010update}. 

Many existing large-scale 3D city models, particularly those at a Level of Detail 2 (LOD2), are primarily derived from aerial-based methods such as Airborne Laser Scanning (ALS) or photogrammetry \citep{wysocki2022refinement,aringer2014bavarian}. While these methods provide excellent coverage of rooftops and general building footprints, they struggle to capture detailed facade information, which is often occluded or sampled at very low resolution. Consequently, the resulting facade geometries often deviate significantly from the actual, real-world structures.

To overcome these limitations, research has shifted towards multi-source data fusion, such as combining ALS with Mobile Laser Scanning (MLS) or photogrammetry \citep{xu2018pointfusion, tysiac2023combination,li2024adaptive}. However, reconstructing models from scratch using multi-source data introduces significant challenges. The registration of such heterogeneous data sources remains a complex and error-prone task \citep{huang2023cross}. Furthermore, different datasets are often acquired at different times, leading to temporal inconsistencies and inaccuracies \citep{xu2024high}. Most importantly, the high computational and resource costs of a ``from scratch'' reconstruction make it impractical for the frequent updates required by dynamic urban environments \citep{farshian2023deep}.

A more practical and efficient approach is to refine existing models rather than completely reconstructing them. Addressing this need, we propose a novel method for refining 3D building models that leverages the overall structure and topological relationships of the existing coarse model as a baseline and integrates newly measured high-resolution data (\eg, MLS point clouds) to enhance accuracy and detail. The main contributions of this paper are as follows:

\begin{itemize}
    \item A multi-source fusion method for refining existing 3D building models using high-resolution point clouds, specifically correcting inaccurate or outdated facade geometries.
    
    \item An extension of the PolyFit~\cite{nan2017polyfit} framework from full-scale reconstruction to targeted model refinement. By utilizing the coarse model as a geometric prior, this approach guarantees strictly manifold and watertight outputs.
    
    \item A confidence-driven mechanism integrated into the binary integer optimization, which explicitly quantifies the geometric alignment between generated candidate faces and the multi-source inputs.
\end{itemize}

\section{Related Work}\label{relatedwork}

Large-scale 3D city modeling traditionally relies on airborne sensing to generate LOD2 models \citep{xu2021toward,wysocki2022refinement}. For from-scratch reconstruction, early continuous mesh generation methods \citep{hoppe1992surface,bernardini2002ball, kazhdan2006poisson} often over-smoothed sharp architectural features. Primitive-based approaches \citep{lafarge2012creating,lin2013semantic, verdie2015lod,li2022ransac} address this by extracting geometric shapes via RANSAC \citep{schnabel2007efficient}, but frequently fail to guarantee topological consistency, leading to gaps or self-intersections. 

To enforce valid topologies, PolyFit \citep{nan2017polyfit} frames reconstruction as an optimal face selection problem via integer linear programming, which guarantees strictly manifold and watertight outputs. Despite its geometric precision, it suffers from high computational complexity and sensitivity to incomplete data. To improve scalability for large-scale urban scenes, \citet{bauchet2020kinetic} introduced a kinetic partitioning scheme that replaces static global slicing with dynamic space decomposition. Nevertheless, such purely geometric methods remain highly sensitive to data incompleteness. More recently, learning-based approaches \citep{mescheder2019occupancy,park2019deepsdf,chen2022reconstructing,chen2025parametric} infer geometries directly from noisy data, but often struggle with over-smoothing or sensitivity to raw data quality. Given the difficulty of balancing computational cost, data incompleteness, and geometric accuracy in from-scratch reconstruction, refining existing coarse models has emerged as a practical alternative.

Refinement techniques enhance existing coarse models using supplementary information. Early methods fused airborne LiDAR with optical imagery \citep{becker2007refinement} or focused on feature-preserving mesh decimation \citep{li2021feature}, which assumes structural completeness. To reconstruct missing geometries, some approaches target specific structures, such as using CSG boolean operations to carve underpasses \citep{wysocki2022refinement}, but lack a global optimization framework. 

For global geometric repair, \citet{yu2022repairing} proposed a kinetic data structure that decomposes 3D space by extending the facets of the imperfect input model, successfully fixing self-intersections and gaps to guarantee watertight and manifold outputs. However, as a pure mesh repair technique, it operates solely on the existing imperfect facets and cannot fuse new sensory data to update outdated or inaccurate geometries.

To integrate new measurements, recent works fuse MLS data with existing meshes. Data-driven frameworks like MS3DQE-Net \citep{liu20233d} improve facade completeness but are computationally intensive and lack watertight guarantees. Alternatively, multi-modal approaches like Scan2LoD3 \citep{wysocki2023scan2lod3} refine models via Bayesian networks, but strictly depend on sensor trajectories and precise base model pre-alignment.

Despite these advancements, a critical gap remains. Existing watertight reconstructions rely on heavy semantic inference or predefined primitives, limiting geometric fidelity to actual point clouds. Conversely, data-driven fusion and pure mesh repair methods frequently lack either the capacity to ingest new point clouds for geometric updating or the global topological guarantees. There remains a need for a framework capable of fusing complementary point clouds with existing coarse models within a global optimization, ensuring a watertight, manifold output without relying on strict semantic or trajectory priors.

\section{Method}\label{method}\label{sec:method}

\begin{figure}[h]
\centering
\includegraphics[width=\columnwidth]{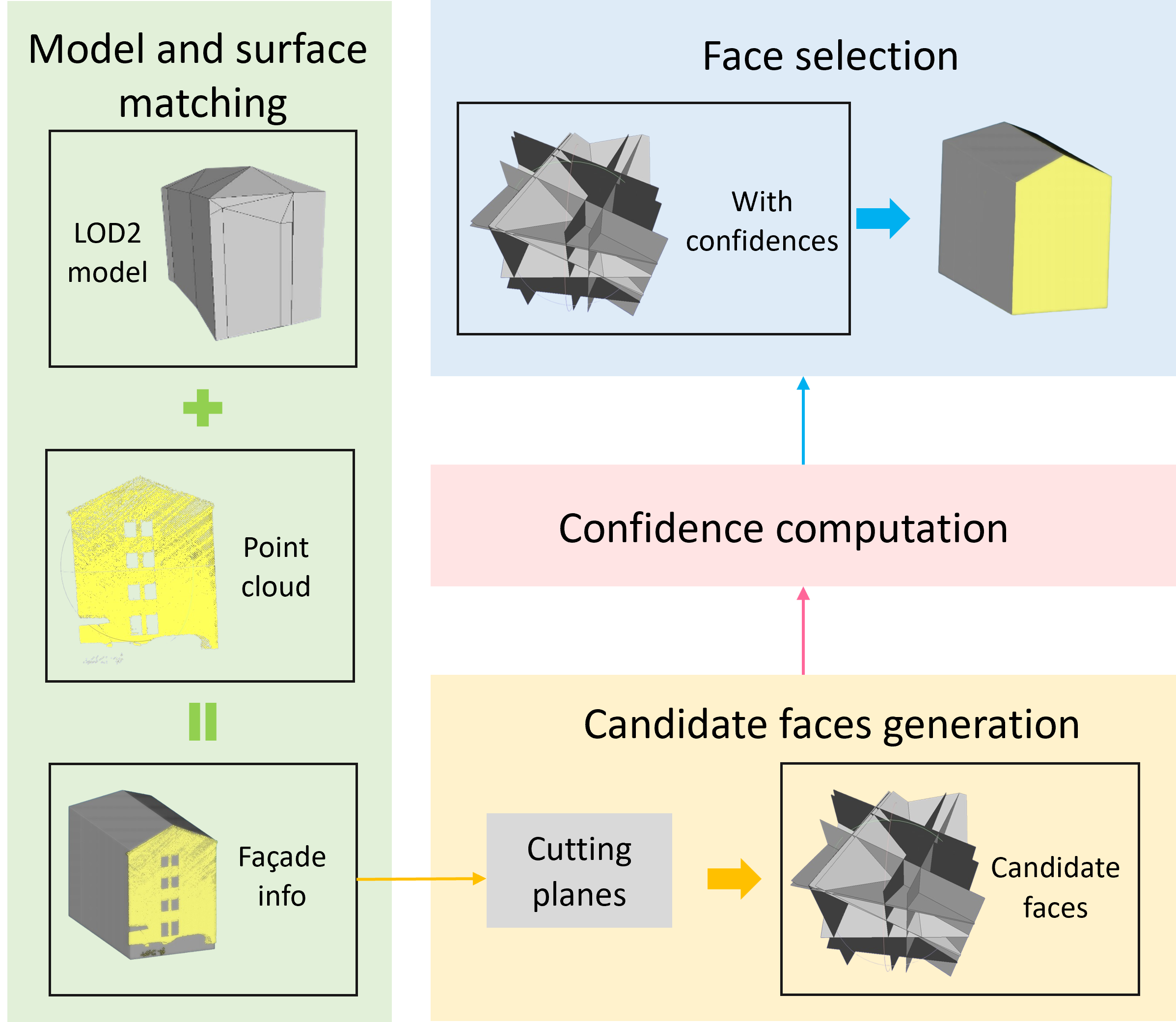}
\caption{Overview of the proposed refinement workflow. The pipeline takes a coarse model and a point cloud as inputs, performing: (1) model to surface matching to extract facade info; (2) candidate face generation; (3) coverage confidence computation; and (4) an optimization-based face selection that guarantees a watertight and manifold refined model.}\label{fig:workflow}
\end{figure}

The proposed method requires two primary inputs: a complete building model, serving as the baseline geometry, and point cloud data for facade updating. The point cloud may cover multiple facades, enabling simultaneous updates to multiple building surfaces.

The refinement workflow is structured into four main components, as shown in Fig.~\ref {fig:workflow}. It begins with model to surface matching, which aligns the building model with the point cloud to identify facades that require updates (step 1). The candidate face generation phase produces a series of potential planar segments within an enlarged bounding box, using the facade information extracted from the matching (step 2). A confidence score is then computed for each candidate face based on its matching information with both the original model and the point cloud (step 3). These confidence scores were used in an optimization process for face selection to generate a watertight building model with accurate updated information (step 4).

\subsection{Model to surface matching}\label{sec:model and surface matching}

Accurately matching the MLS data to the corresponding building models is the critical first step. Given a raw point cloud, we first extract planar primitives using RANSAC \citep{schnabel2007efficient} algorithm. To identify the specific facades requiring updates, we perform a two-stage matching.

First, coarse spatial matching filters candidate building models by intersecting the 2D bounding boxes of the point cloud with those of the expanded CityGML models. Second, a fine-grained geometric matching establishes precise correspondences between the extracted point cloud planes and the building facades through a filtering and scoring mechanism. Initially, a hard filtering step evaluates the normal vector similarity $S_{\text{normal}}$ and the centroid-to-plane distance $d_c$. Pairs failing to meet the predefined normal and distance thresholds are immediately discarded. For the surviving valid candidates, we calculate the spatial overlap of their 3D Oriented Bounding Boxes (OBB), denoted as $C_{\text{bbox}}$ (calculated via volume Intersection-over-Union, IoU).

To rank these surviving candidates and determine the optimal match, a final matching score is computed using a weighted formula that combines the normal vector similarity and the OBB spatial coverage:
\begin{equation}
   S_{\text{match}} = w_{\text{normal}} \cdot |S_{\text{normal}}| + w_{\text{coverage}} \cdot C_{\text{bbox}}
\end{equation}

To handle cases with multiple candidate building models, a model-level quality score is calculated for each model by summing the matching scores of all its filtered matches. The model with the highest total score is then selected as the optimal match for the point cloud. This enables the simultaneous matching of multiple planes and automates the selection of the single best model from a set of candidates. The model-level quality score is defined as:
\begin{equation} 
Q_{\text{model}} = \sum_{k=1}^{N} S_{\text{match}}^{(k)} 
\end{equation}
where $N$ is the number of successfully matched faces within the model, and $S_{\text{match}}^{(k)}$ is the score of the $k$-th matched face.

Once the optimal building model and corresponding facades requiring updates are identified, we remove those outdated surfaces from the original model. Meanwhile, we employ Principal Component Analysis (PCA) \citep{jolliffe2005principal} to compute the planar parameters of the preserved mesh faces in the coarse model. These extracted parameters, along with the bounding boxes of the entire expanded building model (\eg, a 10\% expansion on each side), will serve as core inputs for subsequent steps.

\subsection{Global candidate faces generation}\label{sec:candidate faces generation}

To construct a search space for the optimization, we generate a comprehensive set of global candidate faces. While inspired by PolyFit \citep{nan2017polyfit}, our approach uniquely leverages the matched results from Sec.~\ref{sec:model and surface matching}. Instead of using the entire raw point cloud, our input strictly consists of the matched MLS facade planes, the preserved planar geometries of the original coarse model, and their expanded bounding boxes. This strategy effectively limits the search space to the regions requiring refinement, ensuring we update the existing prior rather than reconstructing from scratch.

\begin{figure}[b]
\centering
\includegraphics[width=\columnwidth]{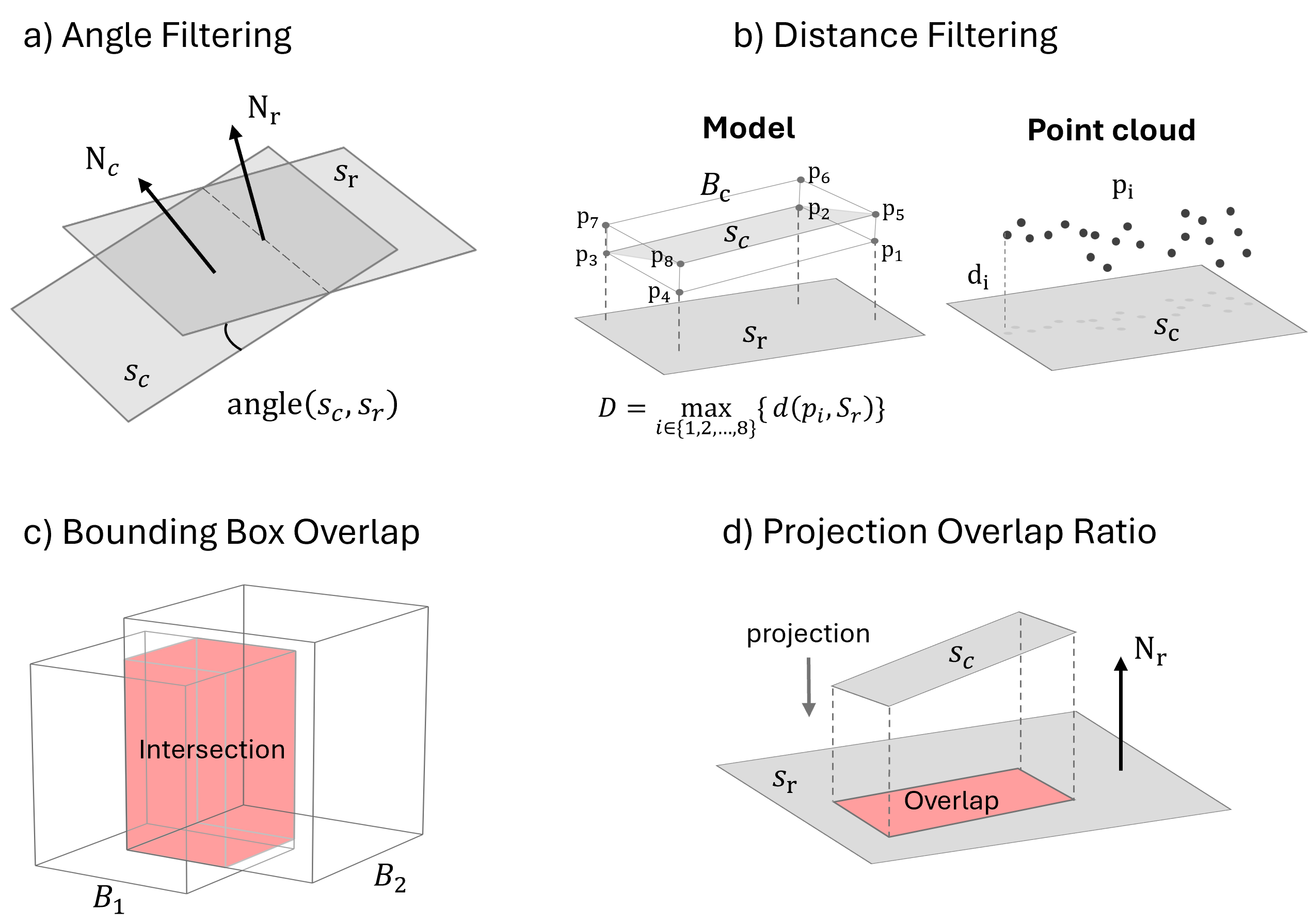}
\caption{Overview of the coverage measuring metrics. The figure illustrates the four key metrics used to evaluate the geometric alignment between a candidate face $s_c$ and a reference plane $s_r$. (a) Angle filtering, which compares the normal vectors $N_c$, $N_r$, (b) Distance filtering, based on the distance $d_i$ of point $p_i$ to the plane, (c) Bounding box overlap, measuring the volume intersection of bounding boxes $B_1$, $B_2$, (d) Projection area overlap, which measures the 2D area of overlap.}\label{fig:coverage}
\end{figure}

We first refine the extracted plane parameters to eliminate geometric artifacts that compromise mesh watertightness, including long, thin faces, small protrusions, and degenerate polygons. Specifically, planar segments with an angular deviation below a threshold $\theta_{merge} = 5^\circ$ are merged into a single plane by averaging their original parameters.

Using these refined supporting planes, we generate a set of global candidate faces through pairwise planar intersections. To restrict the modeling space, all planes are initially cropped by the expanded 3D bounding box. The endpoints of the resulting intersection lines are determined by these cropping bounds, which further define the boundaries of the candidate faces. These candidate faces not only contain geometric information but also preserve adjacency relationships between faces and edges. These relationships are crucial for enforcing manifold and watertight constraints in the final refinement.

\subsection{Confidence Computation}\label{sec:confidence computation}

We compute a confidence score for each generated candidate face, denoted as $s_c$, to quantify its reliability for the final optimization. We base this score primarily on geometric coverage, as the spatial overlap area directly reflects how well $s_c$ is supported by a reference plane $s_r$ from the source data (either the existing coarse model or the MLS points). 

The computation of this confidence score for a given pair $(s_c, s_r)$ follows a sequential filtering pipeline consisting of four metrics (see Fig.~\ref{fig:coverage}). The first three act as hard geometric constraints to filter out invalid candidate-reference pairs, while the fourth computes the final continuous score:

\begin{enumerate}
    \item \textbf{Angle Filtering:} To ensure similar orientations, we discard pairs where the angle between the normal vectors of $s_c$ and $s_r$ exceeds a threshold $\theta_{filter}$. This threshold is typically set to $\theta_{filter} \le 5^\circ$, with the exact value fine-tuned depending on the original model's complexity and level of detail.

    \item \textbf{Distance Filtering:} This metric evaluates geometric proximity. For coarse model references, we filter out pairs if the maximum distance from the bounding box corners of $s_c$ to $s_r$ exceeds a predefined threshold. This threshold is typically set between $0.3$ and $1.0$ meters, depending on the model's overall size and geometric complexity. For point cloud references, we project the points onto $s_c$ and discard the pair if the mean distance or standard deviation of these points exceeds $0.2$ meters, ensuring a tight fit to the dense measurements.

    \item \textbf{Bounding Box Overlap:} Applied exclusively to coarse model facades to ensure spatial alignment, we compute the volume IoU of their bounding boxes. The pair is discarded if the IoU or bidirectional coverage ratio falls below a threshold of $0.3$. This step effectively filters out redundant candidate faces that share similar orientations but lack meaningful spatial overlap with the original geometry.

    \item \textbf{Projection Overlap Ratio (Scoring):} For the surviving valid pairs, the final confidence score is calculated as the 2D projection overlap ratio of $s_c$ onto $s_r$. To ensure robustness against irregular point distributions and noise, the reference area for point clouds is reliably approximated using the 2D convex hull of the projected points.
    
\end{enumerate}

Due to complex geometric relationships such as partial overlaps or overly permissive filtering thresholds, a single candidate face $s_c$ may match multiple reference planes, resulting in multiple valid scores. To ensure the reliability of the final results, we strictly retain only the highest confidence score for each candidate face. This effectively eliminates the influence of low-quality matches, ensuring that the final confidence reflects the optimal geometric correspondence for that candidate face. Candidate faces that fail to find any valid match are assigned a confidence score of $0$.

\subsection{Face selection}\label{sec:face selection}
 Given a candidate faces model composed of $n$ faces, our goal is to identify an optimal subset of $m$ faces ($m \le n$) that accurately captures the new geometry while maintaining structural integrity. We formulate this task as a linear optimization problem, seeking to minimize a composite energy function that contains two key metrics: model coverage and model complexity.

\subsubsection{Model coverage}\label{sec:model coverage}
The coverage term is a critical component of our optimization function, as it quantifies how well the candidate faces represent both the existing model and the newly introduced point cloud data. Since our refinement process operates on an existing model, it is essential to preserve its original geometric characteristics while accurately integrating information from the point cloud. The confidence value, computed for each candidate face in Sec.~\ref{sec:confidence computation}, serves as a measure of its coverage quality.

We integrate this metric into our optimization by defining a cost for each candidate face $i$ based on its area and confidence. The cost, $C_{cov,i}$, is calculated using a sigmoid-based function:
\begin{equation}
C_{cov,i} = A_i \cdot \left( \frac{1}{1 + e^{-(1 - C_i) + (1 - \tau_{cov})}} - 0.5 \right)    
\end{equation}

Here, $A_i$ represents the area of candidate face $i$, while $C_i$ is its confidence value. To control the selection strictness, we introduce a coverage threshold, $\tau_{cov}$. The constant $(1 - \tau_{cov})$ precisely shifts the sigmoid's inflection point to $C_i = \tau_{cov}$. 
Furthermore, subtracting $0.5$ translates the standard sigmoid output range from $(0, 1)$ to $(-0.5, 0.5)$. This zero-centering step guarantees that candidate faces with a confidence $C_i < \tau_{cov}$ (e.g., $\tau_{cov} = 0.3$) incur a positive penalty cost, whereas well-supported faces ($C_i > \tau_{cov}$) receive a negative cost (reward). 
This mechanism encourages the selection of large, highly confident geometric planes. The overall coverage term in our optimization function is the sum of these individual costs for all selected faces, weighted by a parameter $\lambda_{coverage}$.

\subsubsection{Model complexity}
During the global candidate face generation process, as detailed in Sec.~\ref{sec:candidate faces generation}, plane cropping can sometimes produce small, redundant faces. Furthermore, noise and outliers within the point cloud can lead to undesirable artifacts in the final model, such as holes and protrusions. To address these issues and encourage a cleaner, simpler structure, we introduce a complexity term that prioritizes the selection of faces that form large planar regions.

Following the approach of Polyfit \citep{nan2017polyfit}, the term $C_{comp}$ is defined as the ratio of sharp edges in the model. A sharp edge is identified by a pairwise intersection where the two connected faces are not coplanar. By encouraging the selection of coplanar faces, the term promotes the creation of larger, single polygonal regions.
The complexity term is integrated into our optimization function with a configurable weight $\lambda_{complexity}$.

\subsubsection{Linear optimization}
The face selection is formulated as a binary integer programming problem. The objective is to minimize a total cost function, which combines the coverage and complexity terms:
\begin{equation}
\min \quad \lambda_{\rm{coverage}} \sum_{i=1}^n x_i \cdot C_{cov,i} + \lambda_{\rm{complexity}} \cdot C_{\rm{comp}}
\end{equation}
The binary variable $x_i$ is 1 if face $i$ is selected and 0 otherwise. 

Crucially, to ensure the output is strictly manifold and watertight, we enforce hard topological constraints within the optimization solver. Specifically, for every intersection edge $e$ generated in the candidate set, the sum of the binary selection variables $x_i$ of all its incident faces must equal either 2 (if the edge is part of the final surface) or 0 (if the edge is discarded). Mathematically, this is expressed as:
\begin{equation}
\sum_{f_i \in N(e)} x_i \in \{0, 2\}
\end{equation}
where $N(e)$ denotes the set of candidate faces sharing edge $e$. We employ the Gurobi solver \citep{gurobi} to optimize this constrained linear problem. Unselected faces are then deleted, yielding the final closed mesh.

\begin{figure}[t]
  \centering
  \includegraphics[width=0.8\columnwidth]{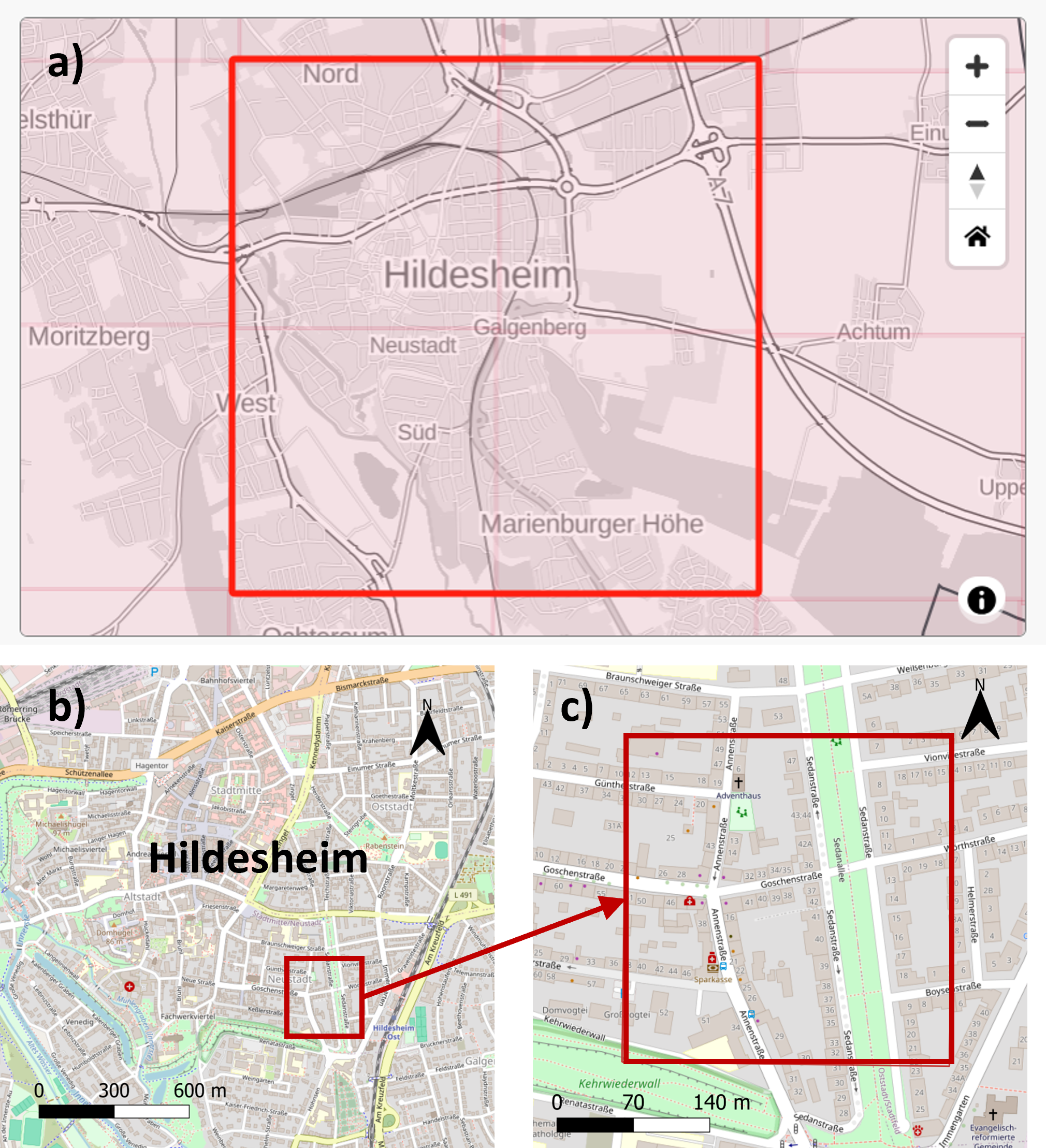}
  \caption{Overview of the study area in Hildesheim, Germany. (a) The spatial extent of the CityGML building models covering the city scale. (b) The geographic context within the city, with the red box indicating the location of the selected test site. (c) Detailed view of the 300 $\times$ 300 m test site used for experiments. (Basemap: OpenStreetMap)}\label{fig:map}
\end{figure}

\section{Experiments and results}
\subsection{Datasets and pre-processing}\label{sec:data}

\begin{figure*}[t!]  
\centering   
\includegraphics[width=0.8\textwidth]{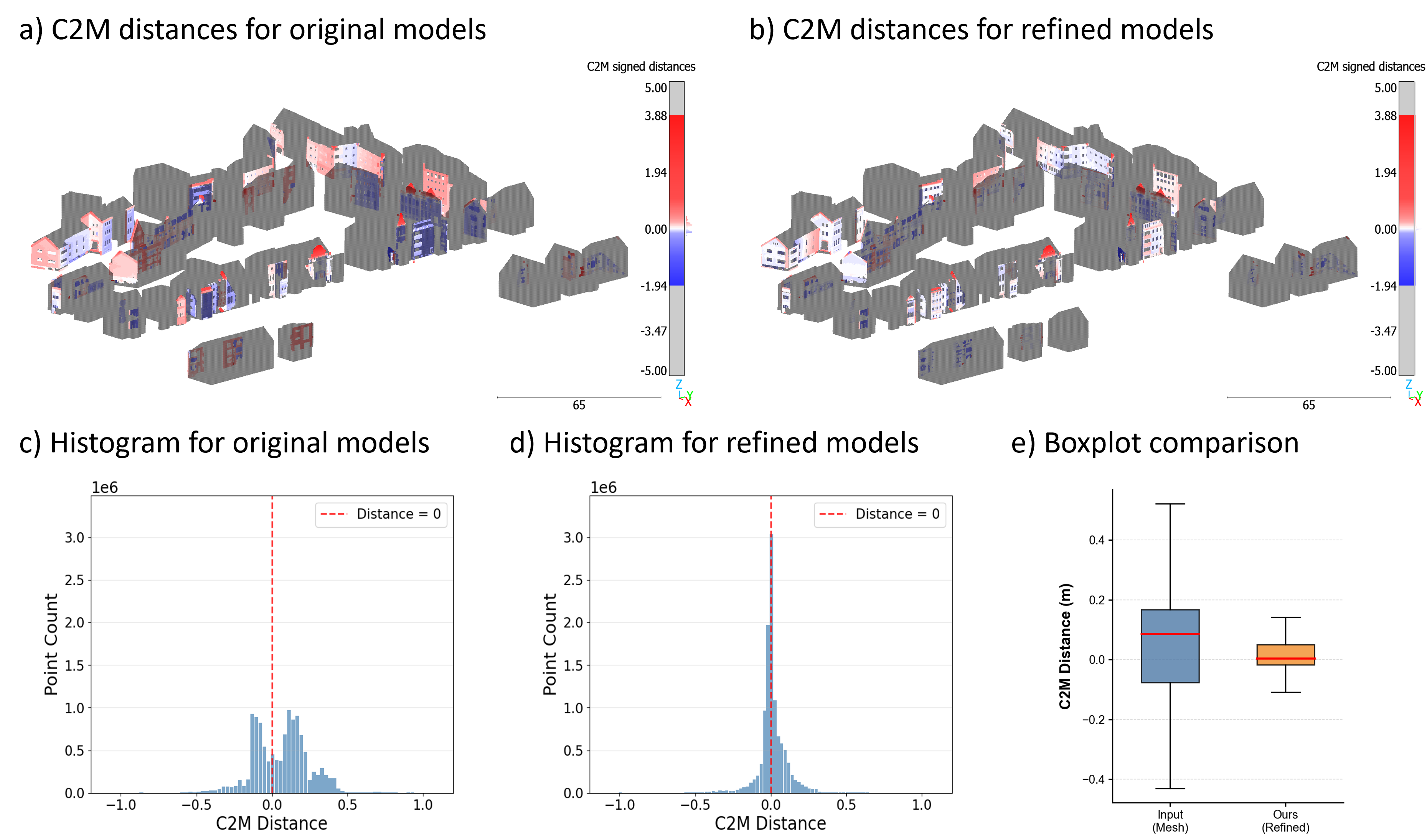}
\caption{Large-scale refinement results on the Hildesheim dataset (approx. 300 m × 300 m). (a) displays the initial coarse models, while (b) shows the refined building models. (c) (d) The original and refined histogram of C2M distances. (e) The boxplot of the distances. The red line represents the median value. The plot compares the error distribution spread of the initial input coarse meshes (blue) versus the refined models (orange).}\label{fig:hildesheim_result1}  
\end{figure*}

The Mobile Laser Scanning (MLS) data used in this work was acquired in Hildesheim, Germany, by \citet{feng2022determination} using a Riegl VMX-250 mobile mapping system. The system consists of two Riegl VQ-250 laser scanners capable of measuring 600,000 points per second \citep{RieglWebsite}. The trajectory was corrected using reference data provided by the SAPOS service \citep{lgln_dgm}. The point cloud exhibits a non-uniform point density ranging from 100 to 50,000 pts/$\text{m}^2$ on building facade surfaces, providing sufficient geometric detail for refinement.

The corresponding coarse models are obtained from the official open geospatial data platform of Lower Saxony (OpenGeoData NI) \citep{niedersachsen_opengeodata}. We retrieved the CityGML LoD2 building models for the Hildesheim region, which geographically align with the acquired MLS data, as shown in Fig.~\ref{fig:map}a.

We selected a specific Region of Interest covering a mixed-use urban block of approximately $300 \times 300$ m for the quantitative evaluation  (see Fig.~\ref{fig:map}c). This experimental dataset comprises 100 individual building models, representing typical building structures with varying complexities. The city-scale dataset was decomposed into individual building objects using an FME attribute-based fanout. Semantic surfaces sharing a common \texttt{gml\_parent\_id} were aggregated and exported as discrete OBJ files keyed by their unique building UUIDs. The MLS point clouds were processed using RANSAC algorithm to extract planar primitives, clustering unstructured points into distinct surface groups.

\subsection{Evaluation metrics}\label{sec:evaluation metrics}
In our quantitative evaluation, the high-density MLS point clouds serve as the geometric reference for evaluating the refinement quality. Deviations from this reference are considered geometric errors. We assess the performance of our refinement framework using the following three metrics:

\begin{enumerate}
    \item \textbf{Geometric Fidelity:} We quantify the geometric accuracy using the Cloud-to-Mesh (C2M) distance, defined as the one-sided Euclidean distance from each point in the reference MLS point cloud to the nearest surface on the refined mesh. We report the Root Mean Square Error (RMSE), the Mean Absolute Distance (MAE), the Signed Mean Distance (Mean) and the Standard Deviation (Std).
    
    \item \textbf{Positional Accuracy:} While C2M distances assess local surface fidelity, they can be insensitive to systematic alignment errors, particularly when obscured by the noise (thickness) of MLS data. We employ the centroid offset reduction ($\Delta D$) to distinguish global positional correction from local shape fidelity.
    Given the spatial discrepancy between the surface-based MLS scans and the volumetric building models, a natural geometric offset exists between their centroids independent of alignment. To isolate the refinement gain, we define $\Delta D$ as the reduction in the absolute centroid distance:
    \begin{equation}
        \Delta D = \|\mathbf{c}_{mls} - \mathbf{c}_{init}\| - \|\mathbf{c}_{mls} - \mathbf{c}_{ref}\|
    \end{equation}
    where $\mathbf{c}_{init}$ and $\mathbf{c}_{ref}$ are the centroids of the initial and refined models. A positive $\Delta D$ confirms that the building volume has been globally translated towards the reference data, verifying the effectiveness of the positional refinement.

    \item \textbf{Topological Validity:} We evaluate the watertightness rate to verify that the refinement eliminates topological errors present in the initial coarse models. We define a model as watertight if it is manifold, containing no holes or open boundary edges. This metric reports the percentage of models that successfully achieve a closed, physically valid surface.
    
\end{enumerate}

\subsection{Parameter sensitivity analysis}\label{sec:parameter_analysis}
\begin{figure}[h]
\centering
\includegraphics[width=\columnwidth]{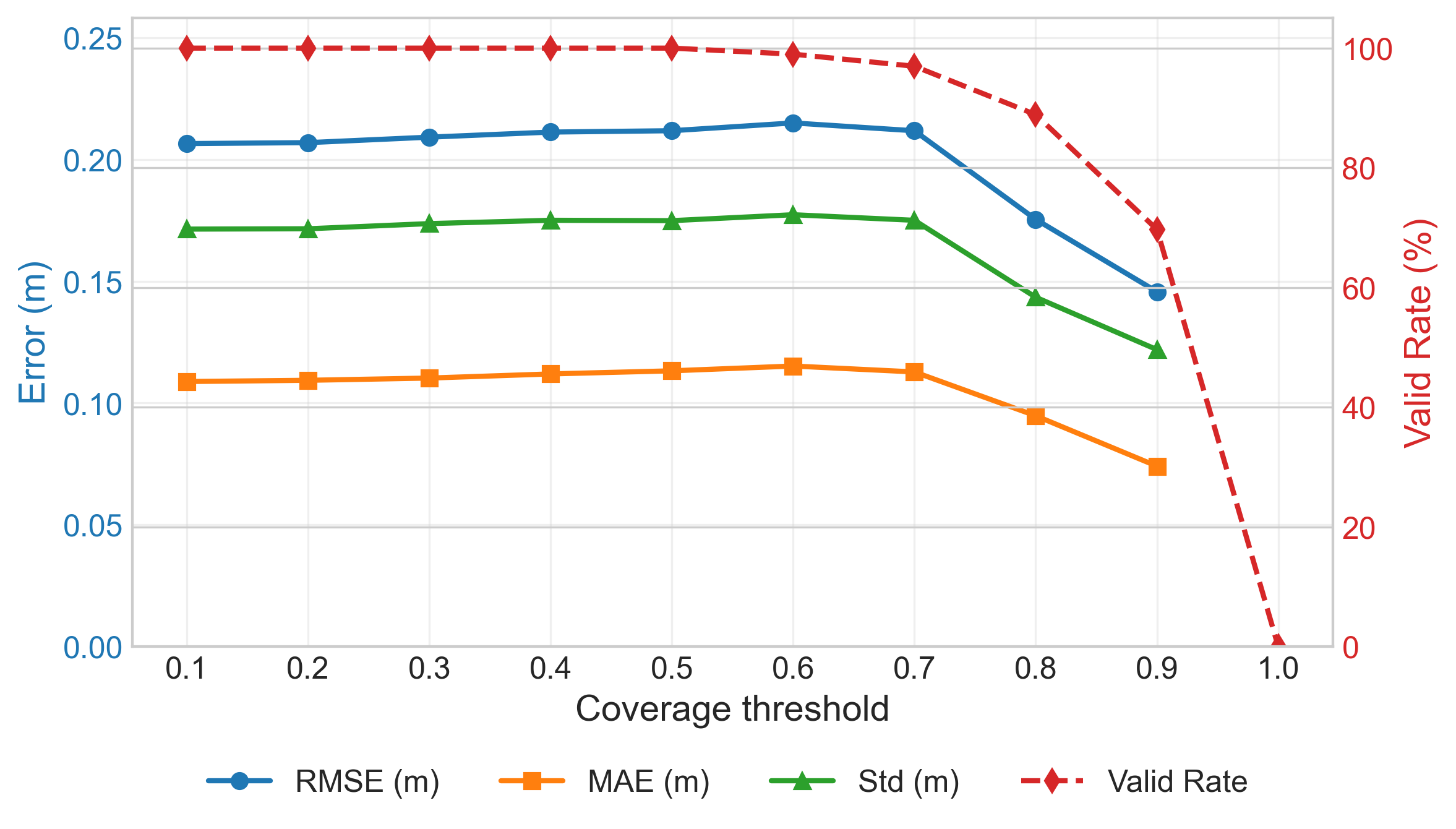}
\caption{Parameter sensitivity analysis on the coverage threshold $\tau_{cov}$. The plot evaluates the average C2M errors (left axis) and the topological validity rate (right axis) across varying thresholds.}\label{fig:parameter_sensitivity}
\end{figure}

Before the large-scale evaluation, we determine the optimal coverage threshold $\tau_{cov}$ (Sec.~\ref{sec:model coverage}) through parameter sensitivity analysis. This analysis was conducted on a representative validation subset comprising 40 typical building instances (40\% of the dataset). We varied $\tau_{cov}$ from $0.1$ to $1.0$ to evaluate the average geometric fidelity (C2M errors) and the topological validity rate (the percentage of successfully generated manifold and watertight models). 
As shown in Fig.~\ref{fig:parameter_sensitivity}, for $\tau_{cov} \in [0.1, 0.5]$, the validity rate remains stable at $100\%$. Within this range, the C2M error metrics also remain consistent (\eg, RMSE $\approx 0.21$ m). 
Conversely, for stricter thresholds ($\tau_{cov} \ge 0.6$), the validity rate drops sharply. Although the average C2M errors decrease, this occurs because a high threshold aggressively rejects candidate faces that lack dense point cloud support or suffer from minor occlusions. This over-filtering prevents the optimization from satisfying the hard manifold constraints, leading to a failure to generate a valid output. 
We empirically set $\tau_{cov} = 0.3$ as the optimal parameter because it provides the strictest possible filtering against noisy candidate faces while safely maintaining a $100\%$ validity rate, thereby achieving the optimal balance between geometric accuracy and valid model generation.

\subsection{Results and analysis}

The quantitative and qualitative performance of the proposed method was evaluated on the Hildesheim dataset using the metrics defined in Sec.~\ref{sec:evaluation metrics}.

Tab.~\ref{tab:watertightness} summarizes the statistical comparison between the original coarse models and the refined results. First, regarding geometric fidelity, the method reduced the C2M distance RMSE from 0.220 m to 0.141 m. Notably, the C2M Mean decreased to 0.016 m, indicating that the deviation of the coarse models relative to the reference MLS point cloud has been effectively eliminated, resulting in a high degree of alignment between the refined geometry and the MLS data. This improvement in geometric accuracy is further corroborated by the error distribution in Fig.~\ref{fig:hildesheim_result1}e. The compressed interquartile range (IQR) and the centering of the distribution around zero in the refined results demonstrate the reduction of error spread and the removal of misalignment. Second, in terms of positional alignment, the centroid offset reduction was calculated at 0.055 m, reflecting the correction of planar shifts. Third, for topological validity, the watertight rate improved from 42\% to 100\%, ensuring that all refined models are geometrically sealed and free of open edges or non-manifold geometry.

\begin{table}[ht]
    \centering
    \renewcommand{\arraystretch}{1.0} 
    \begin{tabularx}{\columnwidth}{Xcc} 
    \toprule 
     & Original model & Refined model \\
    \midrule 
    C2M RMSE (m) & 0.220 & 0.141 \\
    C2M MAE (m) & 0.161 & 0.065 \\
    C2M Mean (m) & 0.067 & 0.016 \\
    C2M Std (m) &  0.210 & 0.140 \\
    Centroid Offset \newline Reduction (m) & - & 0.055 \\
    Watertight Rate (\%) & 42 & 100 \\
    \bottomrule 
    \end{tabularx}
    \caption{Quantitative statistics of the reconstruction quality between the original and refined models.}\label{tab:watertightness}
\end{table}

Fig.~\ref{fig:hildesheim_result1} visualizes the spatial distribution of geometric deviations via C2M signed distance heatmaps and corresponding frequency histograms. The original models exhibit distinct red and blue regions, highlighting the significant misalignment between the coarse building facades and the MLS reference data. This positional discrepancy is quantified by the histogram on the right, which displays a dispersed distribution of distance values. In contrast, the refined results display a predominantly neutral color distribution, while the corresponding histogram shows a sharp peak around zero. This confirms that our proposed method effectively mitigates these positional discrepancies across the large city scale, aligning the model facades with the high-precision measured point clouds.

\begin{figure}[t]
    \centering
    \includegraphics[width=1.0\columnwidth]{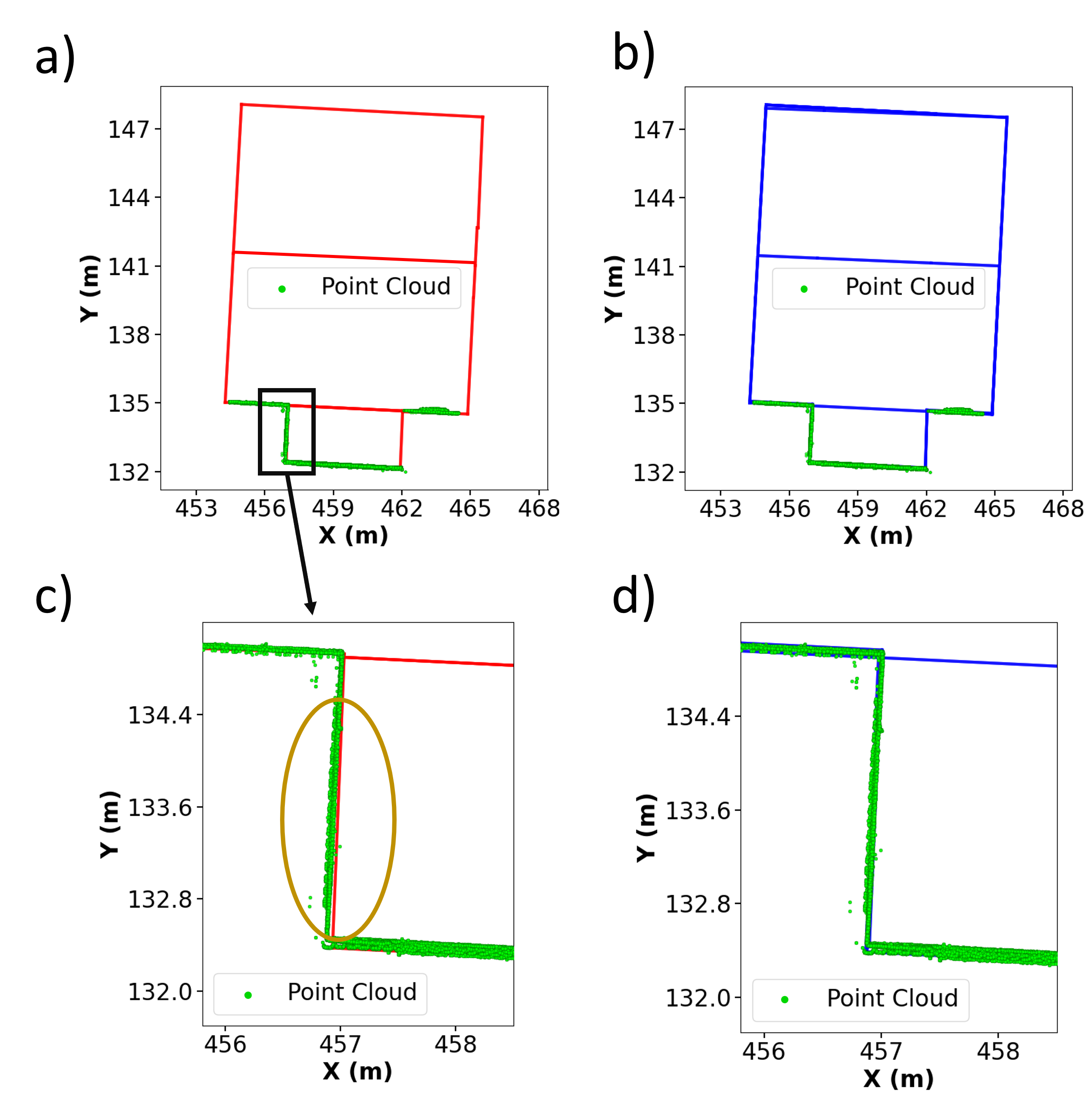}
    \caption{Visual comparison of the building footprint alignment projected onto the 2D XY plane. The global views in (a) and (b) show the original and refined models overlaid on the MLS point cloud (green), where black rectangles indicate the regions detailed in the zoom-in below. As highlighted by the orange ellipse in (c), the original model exhibits visible misalignment, which is effectively corrected in the refined result shown in (d).}
    \label{fig:2d_projection}
\end{figure}

To inspect the alignment fidelity at a finer scale, Fig.~\ref{fig:2d_projection} presents a case study of a representative building footprint projected onto the 2D XY plane. In the global views (Fig.~\ref{fig:2d_projection}a-b), the model is overlaid with MLS point clouds covering four distinct facades, visualized as green linear segments along the building boundary. As illustrated in the detailed view of one specific facade (Fig.~\ref{fig:2d_projection}c), the original model exhibited visible translational and rotational misalignment relative to the scan data. In contrast, the refined result shown in Fig.~\ref{fig:2d_projection}d demonstrates that these boundaries have been corrected to align tightly with the high-density facade point clouds. This visual evidence corroborates the centimeter-level alignment accuracy indicated by the reduced centroid offset and mean error in Table~\ref{tab:watertightness}.

\begin{figure*}[t]
    \centering
    \includegraphics[width=\textwidth]{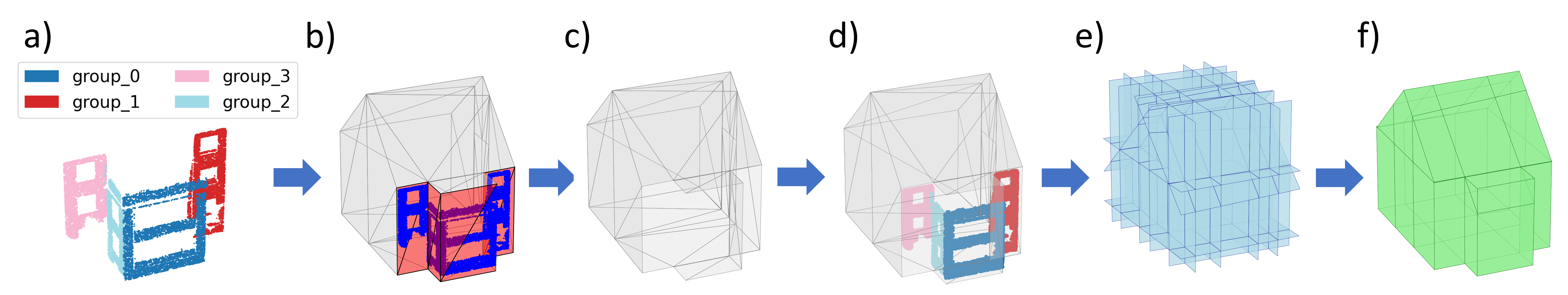}
    \caption{Visual demonstration of the progressive refinement process on a representative building. (a) The input MLS point cloud is segmented into distinct planar groups represented by different colors. (b) The segmented point groups are matched with the corresponding faces of the coarse model via surface matching to identify the facades requiring update. (c) The matched coarse facades are removed from the model. (d) The remaining model structure is integrated with the segmented point clouds. (e) A dense set of candidate planes is generated based on (d). (f) The final watertight model is reconstructed by selecting the optimal faces from (e).}
    \label{fig:process}
\end{figure*}

Finally, Fig.~\ref{fig:process} details the progressive refinement workflow for the same building instance presented in Fig.~\ref{fig:2d_projection}. First, Fig.~\ref{fig:process}a displays the segmentation of raw points into four planar clusters corresponding to the building facades. Guided by these clusters, the surface matching identifies the specific coarse faces to be updated, which are highlighted in red in Fig.~\ref{fig:process}b (matching criteria defined in Sec.~\ref{sec:model and surface matching}). These faces are subsequently removed to form the open mesh shown in Fig.~\ref{fig:process}c. Fig.~\ref{fig:process}d illustrates the fusion of the remaining mesh geometry with the MLS point cloud, which serves as the basis for generating the dense candidate planes in Fig.~\ref{fig:process}e. Finally, the optimization (Section~\ref{sec:face selection}) selects the optimal faces to form the final watertight model in Fig.~\ref{fig:process}f. This workflow verifies the feasibility of reconstructing a valid topology from hybrid input data.

\section{Discussion}

The experimental results demonstrate that the proposed framework effectively integrates high-precision MLS point clouds to refine coarse CityGML models. The significant reduction in C2M RMSE (from 0.220 m to 0.141 m) alongside the suppression of the mean error (0.016 m) indicate that the method successfully corrects the misalignment in the original data. This accuracy improvement is primarily attributed to the utilization of the coarse model as a geometric prior. Unlike reconstruction-from-scratch approaches that typically necessitate complete data coverage and high computational costs, the proposed method performs targeted refinement by constraining the search space with the existing geometry. This strategy not only enhances processing efficiency but also provides a viable workflow for the continuous maintenance of digital twins, where existing models are incrementally updated with newly acquired measuring data.
Furthermore, the optimization enforces topological validity by imposing hard constraints during the face selection process. This guarantees that the output models are manifold and watertight. It eliminates the gaps and disconnected faces often produced by independent reconstruction. However, the refinement accuracy depends on the quality of the input MLS data. The method assumes that the point clouds clearly represent the planar surfaces. In practice, the scan data often exhibits a certain thickness due to sensor noise or reflections from minor facade structures. This thickness introduces uncertainty in the plane estimation. If the data contains significant clutter, the estimated plane position may shift away from the true wall surface. Consequently, the method requires effective pre-processing to filter out such noise and isolate the dominant structural information before the refinement. Additionally, the current assumption of planar primitives may oversimplify complex, non-planar shapes, suggesting a need to incorporate curved surfaces in future work.

\section{Conclusion}

We propose an automated framework for the geometric refinement of coarse LOD2 building models using MLS point clouds. By integrating surface matching with a global optimization strategy, the proposed method addresses the challenges of aligning multi-source data and reconstructing topologically valid geometry.
The quantitative evaluation on the Hildesheim dataset confirms the efficacy of the method. By mitigating the misalignment between coarse models and measured MLS data, the method achieves centimeter-level alignment accuracy, reducing positional deviation and decreasing the C2M RMSE by approximately 36\%. Moreover, the optimization-based face selection guarantees topological consistency, reconstructing strictly manifold and watertight models, thereby eliminating the topological errors present in the original dataset. These results demonstrate that the proposed framework provides a reliable solution for updating digital twins, bridging the gap between coarse building models and high-precision measuring data.

{
    \small
    \bibliographystyle{ieeenat_fullname}
    \bibliography{main}
}

\end{document}